\documentclass[12pt, a4paper]{article}
\usepackage{datetime} 
\usepackage[margin=3cm]{geometry} 
\usepackage{graphicx} 
\usepackage{subfiles} 
\usepackage{natbib}
\usepackage{tipa}
\usepackage{palatino}

\makeatletter 
\def\input@path{{../}} 
\makeatother 
\usepackage{scrextend}
\usepackage{hyperref}
\hypersetup{colorlinks, breaklinks,
            urlcolor=[rgb]{0,0,0.7},
            linkcolor=[rgb]{0,0,0.7},
            citecolor=blue}

\usepackage[labelfont=bf]{caption}
\usepackage{subcaption}
\usepackage{enumitem}
\usepackage{footnote}
\usepackage[hang,flushmargin,bottom]{footmisc} 
\usepackage{tocloft}
\usepackage{ragged2e}

\deffootnote{1.5em}{1em}{\textsuperscript{[\thefootnotemark]}\ }
\deffootnotemark{\textsuperscript{[\thefootnotemark]}}

\usepackage{booktabs}
\makesavenoteenv{tabular}
\makesavenoteenv{table}
\DeclareCaptionLabelSeparator*{spaced}{\\[0ex]}
\setlist{noitemsep}
\usepackage{longtable}

\usepackage{pgfplots}
\pgfplotsset{compat=1.14}
\usepackage{wrapfig}

\newdateformat{monthyeardate}{%
\monthname[\THEMONTH] \THEYEAR}

\newcommand{\HRule}{\rule{\linewidth}{1mm}}

\usepackage{listings}

\makeatletter
\DeclareRobustCommand{\textsupsub}[2]{{%
  \m@th\ensuremath{%
    ^{\hspace{-1pt}\mbox{\fontsize\sf@size\z@#1}}%
    _{\hspace{-1pt}\mbox{\fontsize\sf@size\z@#2}}%
  }%
}}
\makeatother

\usepackage{amsmath}

\usepackage{xcolor, soul}

\begin{document}

\begin{titlepage}

\begin{figure*}[h]
\centering
\includegraphics[width=0.5\linewidth]{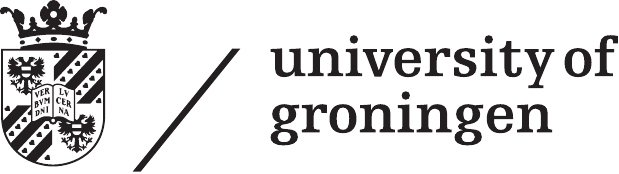} 
\end{figure*}

\begin{figure*}[h]
\centering
\includegraphics[width=0.5\linewidth]{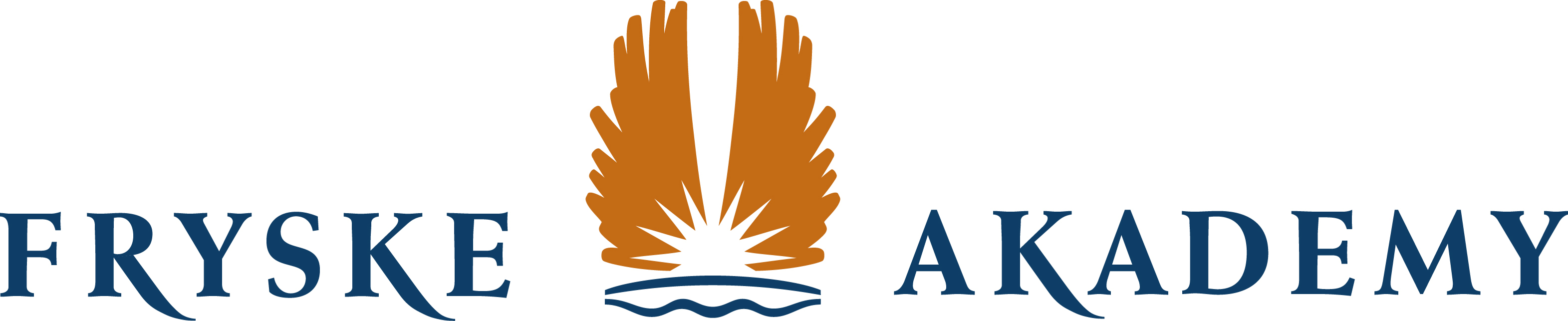} 
\end{figure*}

	\center
	\textsc{\LARGE }\\[3cm]
	
	\HRule \\[0.1cm]
	\Large \textbf{Estimating the Level and Direction of Phonetic Dialect Change in the Northern Netherlands}\\[0.2cm]
	
	\HRule \\
	\Large\monthyeardate\today\\[7.5cm]
	
\begin{table}[!ht]
		\centering
		\begin{tabular}{rl}
			{\textsc{Author}}     & \large{Raoul Buurke}\\
			{\textsc{Author}}     & \large{Hedwig Sekeres}\\
			{\textsc{Author}}     & \large{Wilbert Heeringa}\\
			{\textsc{Author}}     & \large{Remco Knooihuizen}\\
			{\textsc{Author}}     & \large{Martijn Wieling}\\
		\end{tabular}
\end{table}

\end{titlepage}
\newpage

\begin{abstract}
	This article reports ongoing investigations into phonetic change of dialect groups in the northern Netherlandic language area, particularly the Frisian and Low Saxon dialect groups, which are known to differ in vitality. To achieve this, we combine existing phonetically transcribed corpora with dialectometric approaches that allow us to quantify change among older male dialect speakers in a real-time framework. A multidimensional variant of the Levenshtein distance, combined with methods that induce realistic phonetic distances between transcriptions, is used to estimate how much dialect groups have changed between 1990 and 2010, and whether they changed towards Standard Dutch or away from it. Our analyses indicate that language change is a slow process in this geographical area. Moreover, the Frisian and Groningen dialect groups seem to be most stable, while the other Low Saxon varieties (excluding the Groningen dialect group) were shown to be most prone to change. We offer possible explanations for our findings, while we discuss shortcomings of the data and approach in detail, as well as desiderata for future research. 
\end{abstract}

\section{Background}
Dialects in decline show patterns of change, even on linguistic levels that are typically seen as conservative, such as syntax \citep{dorian1973grammatical}. There is no pressing internal need for language varieties with fewer and fewer speakers to undergo significant evolution and change (e.g.,~to update the lexicon with new concepts). However, dialects are typically embedded in a complex language system with political and social influences (cf.~\citealp{auer_2005} for an extensive typology of such systems), and the corresponding dynamics of language contact result in changing varieties. Moreover, younger generations are unlikely to develop their dialect to the same native level as earlier generations due to, e.g., low intergenerational transmission and the absence of a prestige norm given the decline of their dialect. A consequence of this is that the newer generations speak a mixture of, probably neighboring, dialects (see similar cases reported by, e.g.,~\citealp{leopold1959decline, dorian1994varieties}). Numerous studies show that the traditional dialects in the Netherlandic area exhibit patterns of `regiolectization', which is a process that leads to dialects and their neighboring varieties transforming into varieties that occupy an intermediate space between the standard variety and the traditional local dialects \citep{vandekerckhove2009dialect, cornips_2013, swanenberg_2013, wilting2014regiolect}. This new `regiolect' may also stabilize within the linguistic area \citep{ghyselen2015stabilisering} and therefore, at least partially, preserve dialects. 

These patterns and dynamics have so far mainly been highlighted for southern dialect groups in the Netherlands. In this study, we investigate to what degree declining dialects in the northern Netherlandic language area are changing on a phonetic level, and also whether the change we observe is mostly towards Standard Dutch or away from it. Our findings provide some insight into whether the dialects in this geographical area are also candidates for regiolect formation, or are otherwise stable compared to what we know about them from the literature. The Frisian and Low Saxon dialects spoken in this area are of differing vitality  (cf.~\citealp{ytsma2006language, driessen2012ontwikkelingen, bloemhoff_evolution_2013}), which allows us to compare some of the properties of these processes of language change. 

Before we set out to explore these patterns, we provide some background information about the dialects in this geographical area. Specifically, we focus on the Frisian and Low Saxon dialect groups and we leave the dialects of the Low Franconian group out of our analyses (except Standard Dutch, which also belongs to this group). We do so on pragmatic grounds, as this makes it possible for a single transcriber to process all the dialect recordings, but also to avoid having to account for the dynamics \textit{within} the Low Franconian group itself.

\begin{figure}[!ht]
	\centering
	\includegraphics[width=.5\linewidth]{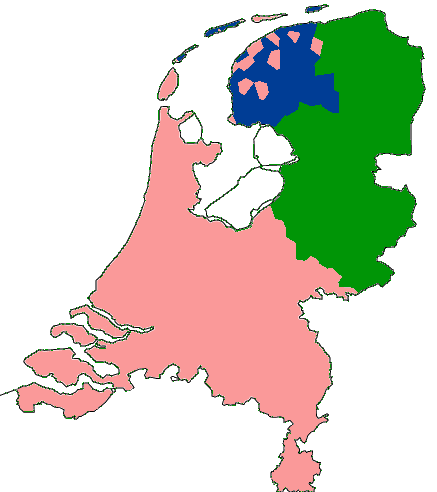}
	\caption{Major language divisions in the Netherlands (largely based on clustering dialects on the basis phonetic similarity in the Goeman-Taeldeman-Van Reenen project data; \citealp{taeldemanfonologie1996}). Pink: Low Franconian. Blue: Frisian. Green: Low Saxon.}
	\label{fig:major_div}
\end{figure}

\begin{figure}[!ht]
	\centering
	\includegraphics[width=.9\linewidth]{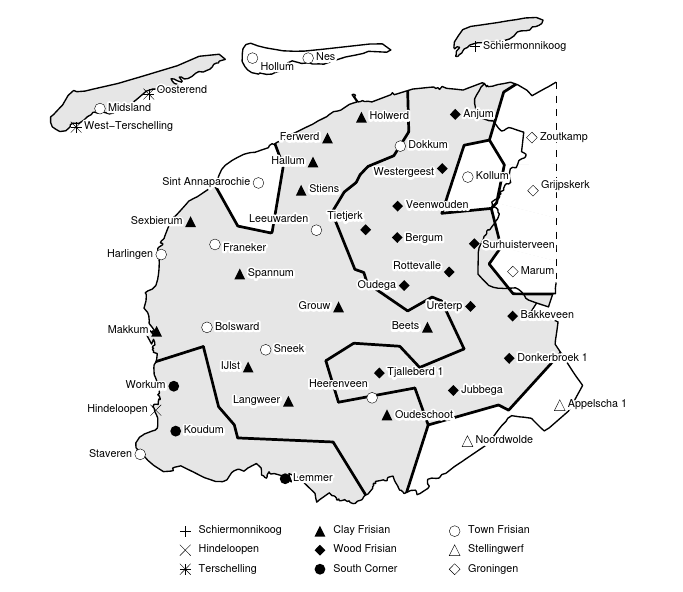}
	\caption{Traditional dialects in Frysl\^{a}n (taken from \citealp{heeringa2005dialect}).}
	\label{fig:frisian}
\end{figure}

The three major language divisions in the Netherlands are shown in Figure \ref{fig:major_div}. Note that the white areas do not concern traditional dialect areas, because they were settled only in the past century due to poldering. The West-Frisian dialect group (referred to here as simply Frisian, because other Frisian varieties are not spoken in the area), spoken in the province of Frysl\^{a}n and shown in more detail in Figure \ref{fig:frisian}, is linguistically distinct from the Low Franconian group and is not readily intelligible to speakers from other varieties in the Netherlands \citep{van2005easy}, because it already split from their common ancestor in the Middle Ages. Nowadays, a distinction is typically made within the Frisian group between Clay Frisian (northwestern area), Wood Frisian (southeastern), and Southwestern Frisian \citep{hoekstra2003frisian}. We expect phonetic differences between these dialects to be particularly salient to Frisian speakers, because they are typically clustered together when non-Frisian data are included as well \citep{nerbonne1999edit}. Moreover, the Frisian dialect group is known to be relatively vital, as it enjoys legal protection under the European Charter for Regional and Minority Languages, and has its own written standard language taught in schools. Furthermore, its speaker population decreases much slower than those of other dialects do \citep{driessen2012ontwikkelingen}. 

We will take one distinction within the province of Frysl\^{a}n into account, however. Specifically, we will distinguish the Dutch-Frisian mixed languages. These dialects include Town Frisian and Bildts (the dialects from around Sint Annaparochie are typically seen as different from Town Frisian; cf. \citealp{duijff2002fries}), visible as the pink spots in Frysl\^{a}n in Figure \ref{fig:major_div}. They have a largely Hollandic lexicon and cluster closely with Dutch, and they are therefore distinct from the other Frisian dialects \citep{gooskens2004position, van2016bildts, van2019fries}. Interestingly, these dialects seem to be relatively conservative \citep{heeringa2000change, versloot2021volatile} despite their  speaker populations decreasing more rapidly than the surrounding Frisian speaker populations.

\begin{figure}[!ht]
	\centering
	\begin{subfigure}{.5\textwidth}
		\centering
		\includegraphics[width=\linewidth]{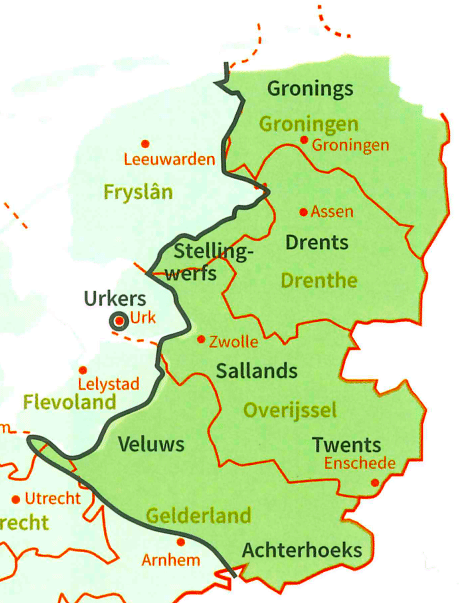}
		\caption{Major dialect groups in the Low Saxon area.}
		\label{fig:ls}
	\end{subfigure}%
	\begin{subfigure}{.5\textwidth}
		\centering
		\includegraphics[width=\linewidth]{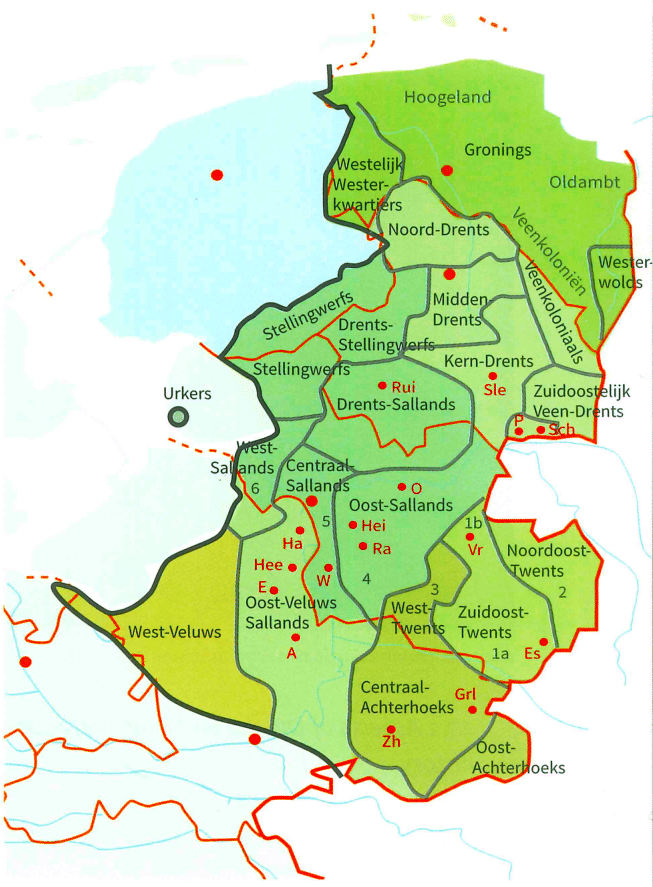}
		\caption{Classification of Low Saxon dialects on the basis of historical linguistic phenomena.}
		\label{fig:groningen}
	\end{subfigure}
	\caption{Dialect maps of the Low Saxon area (adapted from \citealp{bloemhoff2020}).}
	\label{fig:ls_distinction}
\end{figure}

The Low Saxon dialect group, shown in more detail in Figure \ref{fig:ls_distinction}, is also linguistically distinct from the Low Franconian group, though its extensive language contact with Standard Dutch has left its marks on varieties within the group (i.e.,~it is clearly more similar to Standard Dutch than Frisian dialects). As this dialect group covers a much larger geographical area than Frisian, the internal partitioning of the continuum is less easily agreed upon. \cite{bloemhoff2020} made a map (Figure \ref{fig:groningen}) based on the analysis of historically linguistic processes and differences, but the result comprises a large amount of detail that may not be perceivable even by native Low Saxon speakers. We refer to \cite{bloemhoff_low_saxon_phonology_2013} for a detailed description of the main differences between varieties, although they admit that existing classifications of the Low Saxon varieties need further work. 

Within the Low Saxon dialect group, we will explicitly investigate the position of the Groningen dialects, as these constitute a major division in the Low Saxon language family compared to the other dialects when they are computationally clustered \citep{nerbonne2001computational}. Moreover, the Groningen dialects show phonological or morphosyntactic patterns that often overlap with Frisian rather than with other Low Saxon dialects \citep{hinskens2013dutch, van2017frisian}. The Groningen dialects have a substantial Frisian substrate that is absent from the other Low Saxon dialects, as Frisian was only replaced by Low Saxon in Groningen in the late Middle Ages \citep{bloemhoffhandboek2008}, which may explain its greater linguistic distance to the Low Saxon dialects even today. We may therefore expect these dialects to show different patterns from the other Low Saxon dialects in our analyses. Note that we take somewhat broader borders than shown in Figure \ref{fig:groningen}, because we include Westelijk Westerkwartiers, Noord-Drents, Westerwolds, and Veenkoloniaals with the Groningen dialects. This partitioning of Gronings is consistent with studies that phonetically cluster Low Saxon dialects \citep{nerbonne2001computational, heeringameasuring2004}.

There are two main approaches to investigate language change, which depend on the type of data available: analyzing data using the apparent-time construct (see, e.g.,~\citealp{tagliamonte2011variationist}), or analyzing data in real-time. Researchers working within the former paradigm compare speech data from younger generations to that of older ones, which are assumed to reflect newer and older linguistic forms, respectively. The crucial assumption of this paradigm is that individual language systems of speakers remain relatively stable after reaching adulthood. However, this is not always true \citep{blondeau2001real, ashby2001nouveau, sankoff2007language}. If suitable data are available, the preferred approach is therefore a real-time analysis, which samples the language at different points in time among speakers of approximately the same age. Real-time studies are, however, quite costly to conduct \citep{tillery2003approaches}. Luckily, we benefit from the fact that two large dialect data collections have already covered our linguistic area of interest, and we are able to re-use these data. Consequently, we will proceed with a real-time approach in this study. 

In this work, we use a dialectometric approach, which aims to study language varieties (synchronically or diachronically) by aggregating patterns from data across as many locations and words as possible (as opposed to traditional dialectology or dialect geography). Such approaches have been useful for detecting patterns of the Hollandic expansion \citep{kloeke1927hollandsche, wielingquantitative2011}, reliably clustering dialectal varieties \citep{nerbonne1997measuring, heeringameasuring2004, wielingcomparison2007}, and detecting change in dialects due to mutual influence \citep{heeringadialect2015}. Crucially, this approach prevents accidental `cherry picking' of the data, resulting in potentially unreliable patterns. Moreover, we are only fully able to estimate the general rate of (phonetic) change of dialects when we sample a sufficiently large representative portion of its speakers and linguistic system. 

In a preliminary analysis of phonetic language change in the whole Netherlandic area, \cite{buurke2020ma} found that the Low Saxon dialects appeared to have changed more than the Frisian dialects (at least, during a large period in the 20th century). At first, this finding may seem unexpected, because the number of speakers of a Low Saxon dialect is much larger (about two million self-reported speakers; \citealp{bloemhoff_evolution_2013}) than the number of speakers of Frisian (about 470,000; \citealp{provincie_fryslan_fryske_2020}). However, it is in line with \cite{wichmann2009population}, who found that that population size itself is not a significant determinant of language change or stability. In this study, we instead proceed from the assumption that the protective nature of the respective speaker population plays a much larger role in increasing language vitality\footnote{There are conflicting views on what defines language vitality and what contributes to it \citep{mufwene2017language, fitzgerald2017understanding}. When we refer to the vitality of a language variety, we refer to its \textit{usage} in a broad sociopolitical sense and how likely it is to propagated (similarly to the Expanded Graded Intergenerational Disruption Scale; \citealp{lewis2010assessing}). Contributing factors are, i.a.,~language status, policies, economic power, and attitudes of the languages and their speakers in the area \citep{edwards_1992}.} and therefore resistance to standard variety convergence \citep{kristiansen200511}. We know that the Frisian speaker population is much more vital than the Low Saxon one, and so we hypothesize that the rate of change in the Frisian areas is going to be smaller. 

Lastly, as we are not only interested in the amount of change, but also in the direction of change, we look specifically at patterns of convergence and divergence between the dialects and Standard Dutch. \cite{auer2018dialect} labels this type of change `vertical' convergence or divergence in order to distinguish it from processes of neighboring dialects influencing one another (so called `horizontal' convergence or divergence). 

Several studies have shown convergence of dialects in the Netherlands as a whole towards Standard Dutch (for different periods between the 19th and 20th centuries; e.g.,~\citealp{heeringa2000change, heeringaconvergence2014}). However, these studies also note that the patterns are geographically chaotic. Patterns of convergence and divergence may therefore differ substantially between locations that are relatively close to each other. Overall, we expect to find the same general tendency for our data, with convergence contributing more to language change than divergence. However, we focus on a much smaller geographical area than these studies did (i.e.,~Frisian dialects, Dutch-Frisian dialects, Groningen dialects, and the remaining Low Saxon dialects). Specifically, we expect that the Dutch-Frisian and Frisian dialects are relatively stable in our data due to their aforementioned vitality, which in turn protects against extensive influence from Standard Dutch (in the form of convergence). We expect that the Groningen and other Low Saxon dialects converge more towards Standard Dutch, due to their rapidly dwindling speaker numbers and lesser interest in preservation of the dialects of its speakers \citep{bloemhoff_evolution_2013}. 

\section{Data}
A real-time analysis requires multiple samples of the same variety at different points in time. Ideally,  relevant background variables of the speaker samples, such as age, sex, and socioeconomic status, are controlled for. In this study, we use data from two existing phonetically transcribed datasets, the Goeman-Taeldeman-van Reenen project and the From Dialect to Regiolect project. For both datasets, the background of the speakers is relatively similar. 

\subsection{Goeman-Taeldeman-van Reenen project}
The Goeman-Taeldeman-Van Reenen project (henceforth GTRP; \citealp{taeldemanfonologie1996}) was a large-scale undertaking by several academic institutions in the Netherlands and Flanders to collect spoken dialect data across the Netherlandic language area. The data collection for this project took place roughly between 1980 and 1990. Its main aim was to elicit dialect translations of Dutch target words and phrases for dialect geography. This type of data also enables powerful aggregate level analyses that are typically hard to accomplish within the scope of a smaller-scale linguistic study. 

\begin{figure}[!ht]
	\centering
	\includegraphics[width=.95\linewidth]{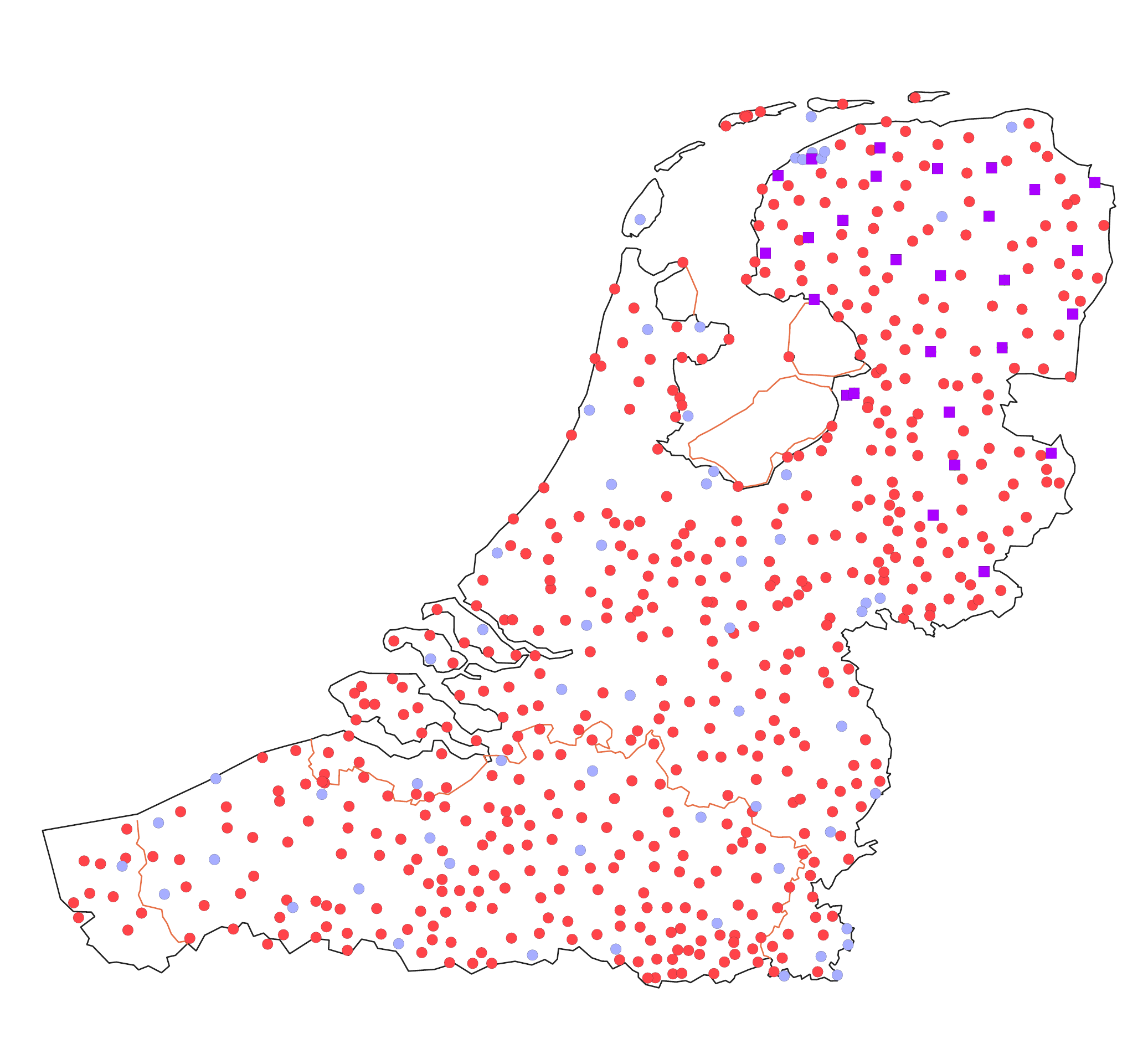}
	\caption{Recording locations. The lighter red and blue dots indicate GTRP and DiaReg locations, respectively. Purple squares (27) are the overlapping locations in the north that are included in this study.}
	\label{fig:locations}
\end{figure}

GTRP field workers went to a total of 613 locations across the Netherlands and Flanders. These locations are shown as red dots (plus the purple squares) in Figure~\ref{fig:locations}. We are limiting the scope of our study to the north of the Netherlands, i.e.,~the provinces of Fryslân, Groningen, Drenthe, Overijssel, and approximately the northern half of Gelderland. We define the southern border of the Low Saxon dialect area at the northern border of the West-Veluws dialect area, which is a transitional area between Low Saxon and Low Franconian \citep{bloemhoffhandboek2008}, with strong tendencies of the latter.

Within the GTRP, participants were selected that ideally were non-mobile older rural males (often abbreviated to NORMs),  in line with what was common in dialectology at the time \citep{chambers1998dialectology}. They were the desired target group, because the archetypal NORM is perceived to be the most conservative in their language use, has undergone little formal education, and is moreover influenced the least by urban dynamics (which induce change through sparse social networks). Speech patterns of these speakers are therefore thought to be reflective of older language forms of their particular space in the dialect continuum. Speakers of the opposite type are mobile younger urban females, who are seen as the most innovative speakers and `drivers' of linguistic change \citep{labov1990intersection, labov1994internal, tagliamonte2011variationist}, though they can definitely be fluent traditional dialect speakers too \citep{goeman2000naast}. These general patterns are consistently found, but there may still be considerable individual variation across particular linguistic variables and sound changes \citep{maclagan_women_1999} or due to personal preferences and attitudes.

GTRP participants were visually presented with the base form of the Standard Dutch target words. They then translated these words into their local variant pronunciation and their pronunciations were recorded and consequently transcribed for the dataset we use for this investigation. The GTRP field workers also elicited morphological variants of verbs and nouns, as there was particular interest in these forms. The total list comprised over 1876 target items, but we only analyze a reduced set of the transcriptions (selected by \citealp{wielingcomparison2007}), because our method of analysis (see Section~\ref{method_change}) is not appropriate for dealing with such variation and would overestimate phonetic change. Forms that are included are therefore base forms of nouns, adjectives, and verbs (the first person plural form). This leaves 562 words per location, which is still a considerable amount of data. 

\subsection{From Dialect to Regiolect project}
The ``From Dialect to Regiolect'' project (DiaReg; \citealp{heeringaconvergence2014}) offers a more recent phonetic dialect data collection from between 2008 and 2011. Its primary aims were slightly different from the GTRP and the composition of the corresponding dataset therefore differs too. One of the aims of the project was to investigate whether there was aggregate-level evidence that dialects in the Netherlandic area were not only converging to Standard Dutch \citep{heeringa2000change, heeringaconvergence2014}, but also towards each other to form so called `regiolects' (a well known possibility in the Netherlandic area and Europe more widely; cf.~\citealp{vandekerckhove2009dialect, swanenberg2011het, cornips_2013, wilting2014regiolect, auer2018dialect}). The DiaReg dataset contains more lexical variation than the GTRP, because the researchers were also interested in lexical change. This was reflected in the design of their experiment, as the researchers elicited running speech, and consequently more lexical variation occurs in the DiaReg dataset. We ensure that these data can be used for our investigation by filtering out realizations that are too distinct for direct comparison (explained in detail further below).
 
Data were collected in a total of 86 locations across the area, which are shown as blue dots (plus the purple squares) in Figure \ref{fig:locations}. Note that the DiaReg researchers aimed for an apparent-time analysis with their data, so they made dialect recordings of both older and younger speakers in each location, which means that their dataset spans the typical sociolinguistic continuum of innovativeness (as mentioned earlier). We use their data in a different manner, however, as we do not compare the older males with the younger females, but we compare the older males from the DiaReg to the older males of the GTRP. 

Participants in this study were presented with a silent movie in the form of stills and (written) narration. The presented story consists of 23 sentences with an average length of 7.6 words, so these were relatively simple sentences. The participants in the study worked in small groups (usually pairs, but up to four speakers in some cases) and each individual was first asked to write down their translated version of the story. Afterwards, they compared their translations and together made a new version everyone agreed with. This `consensus' version was then read aloud by the participants and transcribed. The more involved DiaReg approach avoids two potential pitfalls that typically occur in dialect research: the observer's paradox (i.e.,~noise due to the presence of a researcher; \citealp{labov1972sociolinguistic}) and noise due to differences in individual language systems. We obtained one set of transcriptions per group from the authors, which consisted of a subset of 13 sentences that were consistently pronounced by participants. This subset yields a total of at most 125 target words per location (comprising 90 word types). 

\subsection{Real-time comparison data}
In this study, we use Standard Dutch as a reference point in our analyses. Consequently, we also needed to obtain Standard Dutch target transcriptions. The concept of Standard Dutch pronunciation is notoriously difficult to define, but using the speech of news presenters is seen as a good approximation \citep{smakman2006standard}. Television and radio presenter Maartje van Weegen (around 60 years old at the time) was recorded by \cite{heeringadialect2015} for this purpose. She was tasked with pronouncing the same story as the DiaReg participants did. 

To ensure a real-time comparison between the GTRP and DiaReg with minimal amounts of noise, we limit ourselves to overlapping locations and Standard Dutch target words. There are 27 overlapping locations between the GTRP and the DiaReg (the purple squares in Figure~\ref{fig:locations}). We excluded two locations (Onstwedde and Nijverdal), because the (older male) DiaReg speakers of those locations were born earlier than the corresponding GTRP speakers. Our final dataset therefore contains 25 locations.

Instead of using the pre-existing transcriptions made by DiaReg and GTRP transcribers, the second author transcribed all relevant GTRP and DiaReg recordings anew. This ensures that there is no intertranscriber variability, which is known to be a problem for the GTRP \citep{hinskenspalatalisering2006}. This also allows us to incorporate a greater number of words, as not all target words were transcribed by the GTRP contributors at the time. This procedure yields a maximum overlap of 36 words per location. The overlapping words are summarized in Table~\ref{tab:overlap_words}.

\begin{table}[!ht]
	\caption{Overlapping words between GTRP and DiaReg.}
	\label{tab:overlap_words}
	\begin{tabular}{@{}llllll@{}}
		\toprule
		als    & hem    & is      & loopt & om    & straat  \\
		bij    & hij    & juist   & lopen & ook   & straten \\
		buiten & hoek   & klein   & met   & op    & tijd    \\
		door   & houden & korte   & naar  & ruit  & vraagt  \\
		glas   & huizen & krijgen & niet  & steen & wil     \\
		goed   & in     & later   & nu    & stil  & ziet    \\ \bottomrule
	\end{tabular}
\end{table}

\begin{table}[]
	\caption{Number of available words per comparison after removal of lexical and morphological variation.}
	\label{tab:words_comp}
\begin{tabular}{@{}rrr@{}}
\toprule
\multicolumn{3}{c}{\textbf{Location (number of words)}}              \\ \midrule
Appelscha (22)    & Jubbega (29)    & Sint-Annaparochie (17) \\
Dokkum (21)       & Kampen (32)     & Slochteren (25)        \\
Eelde (28)        & Koekange (33)   & Sneek (16)             \\
Finsterwolde (31) & Laren (23)      & Tilligte (30)          \\
Grijpskerk (26)   & Lemmer (30)     & Veenwouden (31)        \\
Groenlo (32)      & Noordwolde (27) & Workum (29)            \\
Grolloo (32)      & Ommen (25)      & Zwinderen (20)         \\
Grouw (24)        & Roswinkel (28)  &                        \\
IJsselmuiden (32) & Sexbierum (31)  &                        \\ \bottomrule
\end{tabular}
\end{table}

We have to account for lexical and morphological variation in the data, because the DiaReg dataset was constructed to also investigate change at these linguistic levels. If left unaccounted for, phonetic language change calculated on the basis of these transcriptions would be inaccurate, because morphological or lexical change may drive the apparent change on the phonetic level instead. We therefore manually annotated the transcriptions of the comparisons. If the underlying cognate of a pair of GTRP-DiaReg transcriptions was different, they were left out of analyses. For example, Dutch \textit{steen} `stone' was on occasion translated by a DiaReg dialect speaker into a variant of \textit{kei}, which has roughly the same meaning. Similarly, any variation due to diminutives, conjugation, or phonetically reduced forms (due to strictly morphological variation) was marked and removed as morphological mismatches. The number of words that are used for the comparison after filtering these variants is reported in Table \ref{tab:words_comp}. 

It is immediately clear that there are no pairs of locations for which the full set of 36 words can be used for comparison. This is largely due to the considerable amount of phonetic reduction for a few target words. For example, Dutch \textit{hij} `he' is pronounced as (a variant of the) phonetically reduced form [i] in 68\% of the DiaReg cases. It is possible to replace the Standard Dutch target word with the reduced form, but this would still be problematic, because the GTRP transcriptions then mismatch (GTRP pronunciations are rarely, if ever, reduced, because participants pronounced words in isolation). Consequently, we opted to exclude these data. 

In addition, some locations seem to have noticeably fewer words available for analysis. This applies in particular to Sint-Annaparochie and Sneek, and upon closer inspection this seems to be due to a lack of GTRP transcriptions for these locations. We observe that these speakers have a lexicon that is a unique mix of Dutch and Frisian. For example, \textit{straat} `street' is typically realized as [dik] (cf.~Frisian \textit{dyk}) in these dialects. We observe such Frisian lexemes in the Dia\-Reg transcriptions only (again likely due to the task differences eliciting different kinds of speech). For the GTRP such data of known lexical variants are missing, which suggests that the GTRP transcribers treated pronunciations that did not match the Standard Dutch cognate as missing data. This is therefore not problematically different from our approach, as we would have left out lexically different variants in any case. 

The data for Appelscha were problematic, because upon closer inspection of the data it became clear that the GTRP speaker was clearly a Frisian speaker, while the DiaReg speaker was a Low Saxon speaker. Appelscha is a Frisian-Low Saxon border town and is home to a mixture of the two speaker populations. It turns out that the realizations pertain to the same cognate for 22 GTRP-DiaReg word pairs, so we could in principle include these data, but we chose to err on the side of caution and leave out Appelscha for further analyses nonetheless (yielding a total of 24 included locations). After all these considerations, we are left with approximately 67\% of the data (652 word pairs) for the comparison.

Participant age varies within the same general age group, so we need to look at the distribution of age and recording year to obtain an idea of the time span of language change. This is summarized for the 24 locations in Table \ref{tab:comparison_summ}. It should be noted that not all metadata was available for the GTRP. Specifically, there was no speaker age data available for Grolloo, Ommen, Sneek, and Workum. We observe that the mean age of the DiaReg speakers is slightly higher than that of the GTRP speakers, but not problematically so. Given that the ages for the comparison are roughly the same, we can assume that the time span of language change is approximately equal to the difference in recording years, which is 22 years, on average. 

\begin{table}[!ht]
	\caption{Age and time span summary for the comparison.}
	\label{tab:comparison_summ}
\begin{tabular}{@{}lrr@{}}
\toprule
                     & \textbf{GTRP (mean, std.~dev.)} & \textbf{DiaReg (mean, std.~dev.)}  \\ \midrule
Speaker age          & 61 (8)                & 67 (6)               \\
Recording year       & 1987 (2)              & 2009 (1)             \\
Recording year diff. & 22 (2)                & \multicolumn{1}{l}{} \\ \bottomrule
\end{tabular}
\end{table}

\section{Methods}
\subsection{Levenshtein distance}
The Levenshtein distance is an algorithm that can be used to quantify the difference between two strings \citep{levenshteinbinary1966}. It has been adapted for comparing phonetic strings in dialectometry  \citep{kesslercomputational1995, nerbonne1996phonetic} and historical linguistics \citep{list_potential_2017}. The result of the procedure is a count of how many binary operations are \textit{minimally} necessary to turn one string into another (resulting in and given a particular string alignment). A typical alignment is presented in Table \ref{levenExample}. For each pair of two phonetic segments, it is possible to use one of three operations: an insertion of a segment, a deletion of a segment, or a substitution of two segments by each other. With three substitutions, we have a final Levenshtein distance of 3. However, this alignment is not phonetically sensible, because vowels and consonants are substituted with one another. In order to avoid these alignments, we can set the costs of vowel-consonant substitutions to be artificially high. With this restriction, the optimal alignment becomes different, as is shown in Table \ref{levenExample2}.

\begin{table}[!ht]
	\caption{Levenshtein alignment between dialectal variants of Dutch \textit{straat} (`street'). There are no restrictions on which segments can align to each other.}
	\label{levenExample}
	\begin{tabular}{@{}lcccccc@{}}
		\toprule
		& 1 & 2 & 3 & 4    & 5    & 6    \\ \midrule
		\multicolumn{1}{l|}{String 1}  & s & t & \textscr & o    & d    & \textipa{@}    \\
		\multicolumn{1}{l|}{String 2}  & s & t & \textscr & \textopeno    & \textipa{@}    & t    \\ \midrule
		\multicolumn{1}{l|}{Operation} & - & - & - & sub. & sub. & sub. \\
		\multicolumn{1}{l|}{Cost}      & 0 & 0 & 0 & 1    & 1    & 1    \\ \bottomrule
	\end{tabular}
\end{table}

\begin{table}[!ht]
	\caption{Phonetically sensible Levenshtein alignment between dialectal variants of Dutch \textit{straat} (`street'). Here consonants and vowels are not allowed to align to each other.}
	\label{levenExample2}
	\begin{tabular}{l|ccccccc}
		\toprule
		& 1 & 2 & 3 & 4    & 5    & 6    & 7    \\ \hline
		String 1  & s & t & \textscr & o    &      & d    & \textipa{@}    \\
		String 2  & s & t & \textscr & \textopeno    & \textipa{@}    & t    &      \\ \hline
		Operation & - & - & - & sub. & ins. & sub. & del. \\
		Cost      & 0 & 0 & 0 & 1    & 1    & 1    & 1   \\ \bottomrule
	\end{tabular}
\end{table} 

In this second alignment the Levenshtein distance has increased by one, but the alignment itself is linguistically more sensible. Note that we do allow [\textipa{@}] to be aligned with the sonorants [m, l, n, r, \textipa{\ng}, j, w], because sonorant consonants have noticeably more acoustic energy than other consonants, are typically also voiced, and they can be found in vowel positions, such as in the synchronic variation of Dutch \textit{vier} `four', [fi\textlengthmark r] vs.~[fi\textlengthmark \textipa{@}] (p. 125, \citealp{heeringameasuring2004}). After determining the Levenshtein distance, we normalize it by dividing the distance by the length of the longest optimal alignment (in line with \citealp{heeringameasuring2004}, p.~131). If we assume the alignment in Table \ref{levenExample2} is the single optimal alignment, the normalized Levenshtein distance is then 4/7 ($\approx$ 0.57). 

A final optimization of the Levenshtein algorithm is to use linguistically sensitive costs for the operations instead of the binary weights we have used so far in the examples. This approach ensures that a substitution of phonetically more dissimilar segments (e.g.,~[i]-[u]) is assigned a higher cost compared to a phonetically similar segments (e.g.,~[i]-[\textsci]). In order to obtain these sensitive costs, we use the (point-wise mutual information-based, PMI) procedure proposed by \cite{wieling2012inducing}, which is a refinement of earlier efforts \citep{ristad1998learning, wielingevaluating2009}. In this procedure, costs for operations involving specific sounds are induced from their co-occurrence patterns in a particular dataset with phonetic transcriptions (as long as it is of a sufficient size). Consequently, sounds which co-occur more frequently in the alignments are assigned lower substitution costs (closer to 0) than those which co-occur only infrequently (closer to 1). We refer to \cite{wieling2012inducing} for the exact details of the algorithm. Importantly, this approach is known to correlate well with perceptual distances from speakers \citep{wielingmeasuring2014}.

\subsection{Measuring the direction of change} \label{method_change}
In addition to the amount of language change, we also analyze the direction of change by employing a three-dimensional (3D) version of the aforementioned Levenshtein distance. The underlying principles of the two-dimensional Levenshtein distance (i.e.,~for comparing two strings) can be extended to higher dimensions as well \citep{heeringadialect2015}. Using a three-dimensional version of the algorithm, we can compute the difference between strings, while taking another string as a reference point, such as a transcription of the standard language variant. An example is given in Table \ref{tab:leven_3d}, with binary costs for illustration. 

\begin{table}[!ht]
	\caption{3D Levenshtein alignment between dialectal variations of Dutch `straat'.}
	\label{tab:leven_3d}
\begin{tabular}{@{}lccccccc@{}}
	\toprule
	& 1 & 2 & 3 & 4      & 5    & 6     & 7     \\ \midrule
	\multicolumn{1}{l|}{Older variant}            & s & t & \textscr & o      &      & d     & \textipa{@}     \\
	\multicolumn{1}{l|}{Newer variant}            & s & t & \textscr & \textopeno      & \textipa{@}    & t     &       \\
	\multicolumn{1}{l|}{Standard variant}         & s & t & \textscr & a      &      & t     &       \\ \midrule
	\multicolumn{1}{l|}{Older-standard operation} & - & - & - & subs.  & -    & subs. & del.  \\
	\multicolumn{1}{l|}{Older-standard cost}      & 0 & 0 & 0 & 1      & 0    & 1     & 1     \\ \midrule
	\multicolumn{1}{l|}{Newer-standard operation} & - & - & - & subs.  & del. & -     & -     \\
	\multicolumn{1}{l|}{Newer-standard cost}      & 0 & 0 & 0 & 1      & 1    & 0     & 0     \\ \midrule
	\multicolumn{1}{l|}{Direction of change}     & - & - & - & neutr. & div. & conv. & conv. \\ \bottomrule
\end{tabular}
\end{table}

Note there are now seven possible operations instead of three and each operation constitutes a change in all strings simultaneously: inserting a segment from one of the three strings (i.e.,~deleting a segment in the other two: three operations), and substituting segments between each possible pair of strings (e.g.,~between string 1 and 2, 1 and 3, and 2 and 3: three operations), as well as substituting a segment from each of the three strings at the same time (one operation) (cf.~\citealp{heeringadialect2015}, p.~26). 

From this single 3D alignment (for which we have used binary weights to facilitate interpretation), we can obtain the distances between the older variant and the standard, and between the newer variant and the standard.\footnote{Note that the distance between the older and newer variant can also be obtained from this alignment, but we are not interested in this particular distance for our analysis.} For each segment of this alignment, we can then determine whether it reflects a pattern of convergence, divergence, neutral change, or stability. A particular segment is convergent if the distance to the standard sound is greater for the sound in the \textit{older} variant than for the sound in the \textit{newer} variant. If instead the distance to the standard is smaller for the sound in the \textit{older} variant, then the segment concerns a divergent pattern. If these distances are equal, then this segment concerns either neutral change or stability (depending on whether the sounds are equal). 

\cite{heeringadialect2015} determined the type of change for each segment in a categorical manner, but we do so in a gradual manner. We account for the amount of change for each segment by using the PMI weights. This makes the possibility of observing neutral change less likely, because each distance between symbols has a gradual value between 0 and 1. Neutral change can only occur if different sounds occur with the exact same frequency in the transcriptions on which the PMI distances are based. For segment 4 in the example alignment, this would mean that the direction of change is either convergent (if the cost of [o]-[a] is greater than [\textopeno]-[a]) or divergent (if the cost of [o]-[a] is less than [\textopeno]-[a]). We therefore use the following formula to determine the direction of change for each segment:

\begin{equation}
	\label{seg_direction}
	\text{direction}_\text{(x, y, z)} = \text{distance(y, z)} - \text{distance(x, z)}
\end{equation}\\

Note that it is important to ensure that phonetic transcriptions are always compared in the same order, because otherwise the interpretation differs. Following the above equation, we always make sure that \textit{x} is the older variant (the GTRP transcription) and that \textit{y} is the newer variant (the DiaReg transcription), while \textit{z} is the standard variant. When this order is kept, a value  greater than 0 indicates divergence for a segment and a value smaller than 0 indicates convergence. For each 3D alignment, we consequently sum PMI costs for convergent and divergent costs separately, which we then divide over the alignment length. The result of the procedure is then (1) a proportion of convergence and (2) a proportion of divergence for each alignment.

We prefer the alignments generated by this 3D procedure over pairs of alignments generated by the typical two-dimensional (2D) procedure, because they better match our goal of investigating diachronic language change rather than synchronic variation.  In principle, it is possible to reach the `same' goal by generating the typical 2D alignments between the GTRP--standard and between the DiaReg--standard transcriptions and then simply subtracting these two values. Indeed, the distances obtained by double-2D and single-3D alignment approaches are highly correlated ($r > 0.95$, $p < 0.01$). The 3D alignment, however, is conceptually better suited to investigating diachronic change, because a 3D alignment generates more possible segments of change. For example, let us consider the case in Table \ref{tab:leven_3d}. If we analyzed these transcriptions using two 2D alignments, the normalized distance to the standard for the newer variant would be slightly larger than if we used a single 3D alignment (because there is one more empty segment in the alignment). We know that there was actually a segment after the final consonant ([d/t]) for the older variant, which was occupied by a [\textipa{@}] at the time. We would ignore this historical property of the cognate if we did not use a 3D alignment, which is undesirable. The 3D version is therefore more suited than the 2D version to estimating diachronic differences.

\subsubsection{Statistical analysis}
In our statistical analysis, we model the distributions of language change in two ways: (1) by estimating change due to convergence/divergence on the basis of geographical coordinates, and (2) by estimating change due to convergence/divergence for known dialect groups separately. The geographical type of analysis is useful for investigating whether convergence and divergence patterns are spatially gradual (e.g.,~whether more southern dialects show more change than northern ones), and whether convergence and divergence levels differ at large. With the group-based analysis we highlight differences in change between dialect groups, which we expect to behave differently on the basis of their linguistic history (e.g.,~we separate the Dutch-Frisian and Groningen dialects from the Frisian and Low Saxon groups).

For both analyses, we fit generalized additive mixed models (GAMM; \citealp{wood2017generalized}), which also allows us to specify an appropriate random-effects structure. This enables us to obtain reliable results despite the different amount of transcription data available per recording location. We evaluate a successively more complex model using a step-wise model comparison procedure. A particular predictor is added to the model if this more complex model offers a significant improvement over the model without the predictor (cf.~\citealp{wieling_analyzing_2018}).

In order to distinguish convergence and divergence in the data, we follow a similar procedure as \cite{heeringadialect2015}, summing the costs of convergent and divergent segments \textit{separately} for each alignment. In contrast to \cite{heeringadialect2015}, however, we use PMI weights instead of acoustically measured weights, so we end up with a separate value of convergence and divergence for each alignment. We normalize these values using the alignment length, so that we obtain a proportion of each type of change for each alignment, which serves as the dependent variable in our analyses (i.e.,~a value of 1 is the theoretical maximum value). Convergent and divergent change are analyzed in a single model, by including a binary predictor variable distinguishing between the two types for each data point.

The first analysis investigates general patterns of convergent and divergent change across the geographical area of interest. In this analysis, geography is included as a predictor by representing it as a non-linear smooth interaction between longitude and latitude (see \citealp{wielingquantitative2011} for more details about this approach). 

For the second analysis, we replace the geographical smooth with a predictor distinguishing the four dialect groups (Frisian: FR, Dutch-Frisian: DU-FR, Groningen: GR, and the other Low Saxon dialects: LS). The locations per dialect group are summarized in Table \ref{tab:dialectgroup}. 

\begin{table}[!ht]
\caption{The manually selected locations constituting each dialect group.}
\label{tab:dialectgroup}
\begin{tabular}{@{}ll@{}}
\toprule
\textbf{Dialect group (n locations)} &                                  \\ \midrule
FR (7)                      & Grouw, Jubbega, Lemmer,          \\
                            & Noordwolde, Sexbierum,           \\
                            & Veenwouden, Workum               \\
DU-FR (3)                   & Dokkum, Sint Annaparochie, Sneek \\
GR (5)                      & Eelde, Finsterwolde, Grijpskerk, \\
                            & Roswinkel, Slochteren            \\
LS (9)                      & Groenlo, Grolloo, IJsselmuiden,  \\
                            & Kampen, Koekange, Laren,         \\
                            & Ommen, Tilligte, and Zwinderen   \\ \bottomrule
\end{tabular}
\end{table}

\section{Results}
\subsection{Geographical pattern of change}
\begin{table}[!ht]
\caption{Parametric coefficients for a GAMM predicting change on the basis of geography.}
\label{tab:parametricGeo}
\begin{tabular}{@{}lrrrrr@{}}
\toprule
                        & \multicolumn{1}{l}{\textbf{Estimate}} & \multicolumn{1}{r}{\textbf{SE}} & \multicolumn{1}{l}{\textbf{$t$-value}} & \multicolumn{1}{l}{\textbf{$p$-value}} & \multicolumn{1}{l}{} \\ \midrule
Intercept (convergence) & 0.015 & 0.003 & 5.347 & \textless{} 0.001 & ***\\
divergence              & 0.005 & 0.003 & 1.449 & 0.148 & \\ \bottomrule
\end{tabular}
\end{table}

\begin{table}[!ht]
\caption{Smooth function terms (including random effects) for a GAMM predicting change on the basis of geography.}
\label{tab:smoothGeo}
\begin{tabular}{@{}lrrrr@{}}
\toprule
                                    & \textbf{Edf} & \textbf{$F$-value} & \textbf{$p$-value} &     \\ \midrule
s(longitude, latitude): convergence & 2.000        & 4.009            & 0.02          & *   \\
s(longitude, latitude): divergence  & 3.128        & 0.709            & 0.61          &     \\
s(word)  & 13.660        & 2.938            & \textless{} 0.01          & **     \\
s(word, direction)     & 35.857       & 2.335            & \textless{} 0.001          & ***    \\ \bottomrule
\end{tabular}
\end{table}

\begin{figure}[!ht]
	\centering
	\includegraphics[width=.95\linewidth]{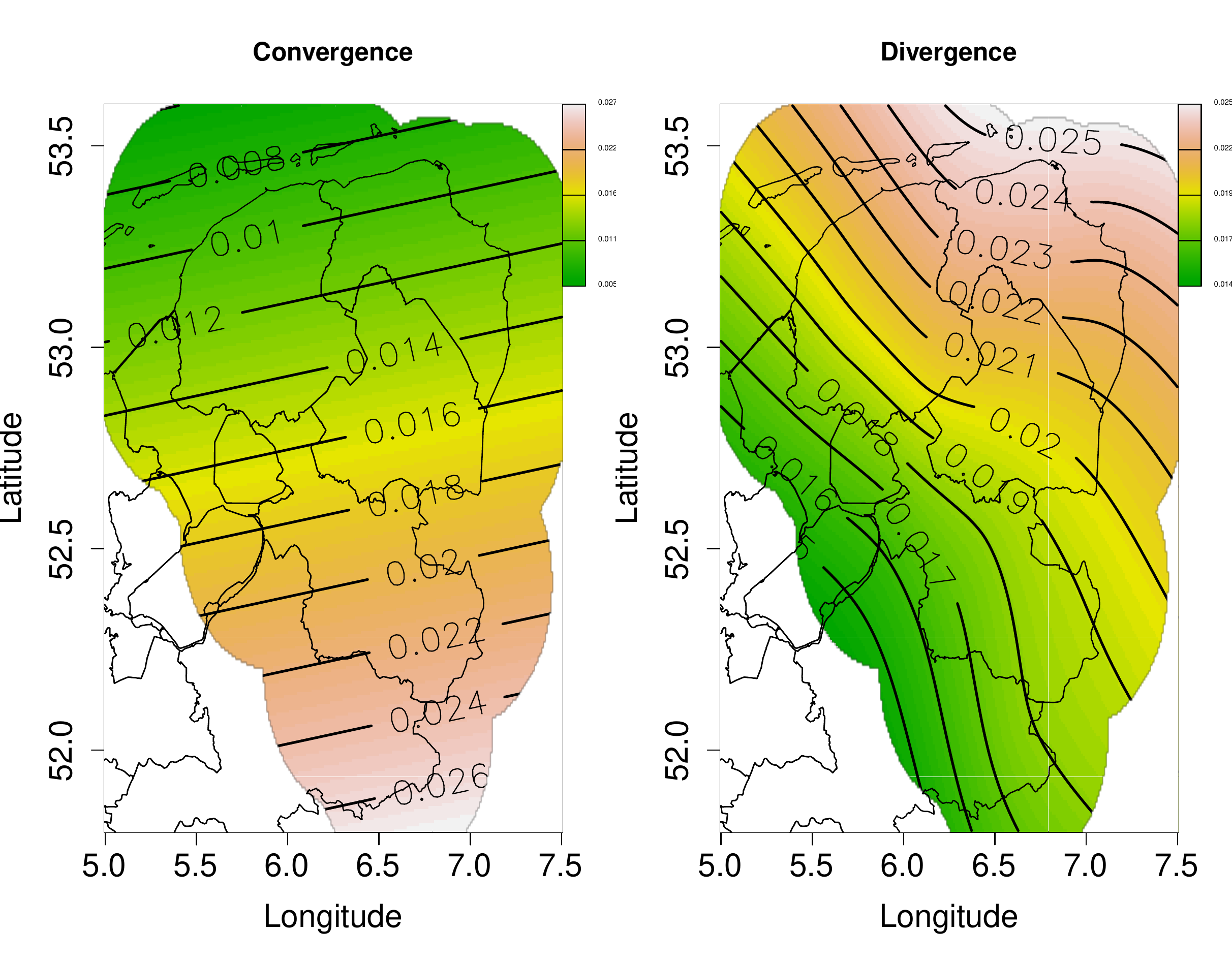}
	\caption{Phonetic change distribution across the geographical area (greener = less change; orange-white = more change). Left: convergence to Standard Dutch. Right: divergence from Standard Dutch. Note that the divergence pattern is not significantly different from zero ($p = 0.61$).}
	\label{fig:change_om}
\end{figure}

The results for the first analysis are summarized in Figure \ref{fig:change_om} and Tables \ref{tab:parametricGeo} and \ref{tab:smoothGeo}. They are based on a model that predicts change (separated into convergence and divergence using a binary factor `direction') on the basis of a factor smooth of geography (i.e.,~the non-linear interaction between longitude and latitude) for convergence and divergence. A random intercept for each word and by-direction random slopes for word (all significant) were also included in the model to account for by-item variation. We also tested whether the age and birth year of the GTRP and DiaReg speakers improved the model, but this was not the case, and therefore these variables were left out of the final model. The final model explains 18\% of the variation in the data (i.e.,~the observed patterns of language change). The model specification is:\\ \vspace{1em}

\noindent 
\texttt{normalized\_change $\sim$ direction + s(long, lat, by=direction) +\\ s(word, bs={\textquotesingle}re{\textquotesingle}) + s(word, direction, bs={\textquotesingle}re{\textquotesingle})}\\ \vspace{0.5em}

\noindent
On the basis of the results of this model, we can conclude that there is no significant difference between the overall magnitude of divergence and convergence (i.e.,~the fixed effect of divergence is not significant). However, the geographical smooths are visualized in Figure \ref{fig:change_om}, and show that convergence is greatest in the south of the region and decreases towards the northern areas. The divergent pattern, showing a more or less opposite effect, is not significant, however.

\subsection{Group-based patterns of change}
\begin{figure}[!ht]
	\centering
	\includegraphics[width=\linewidth]{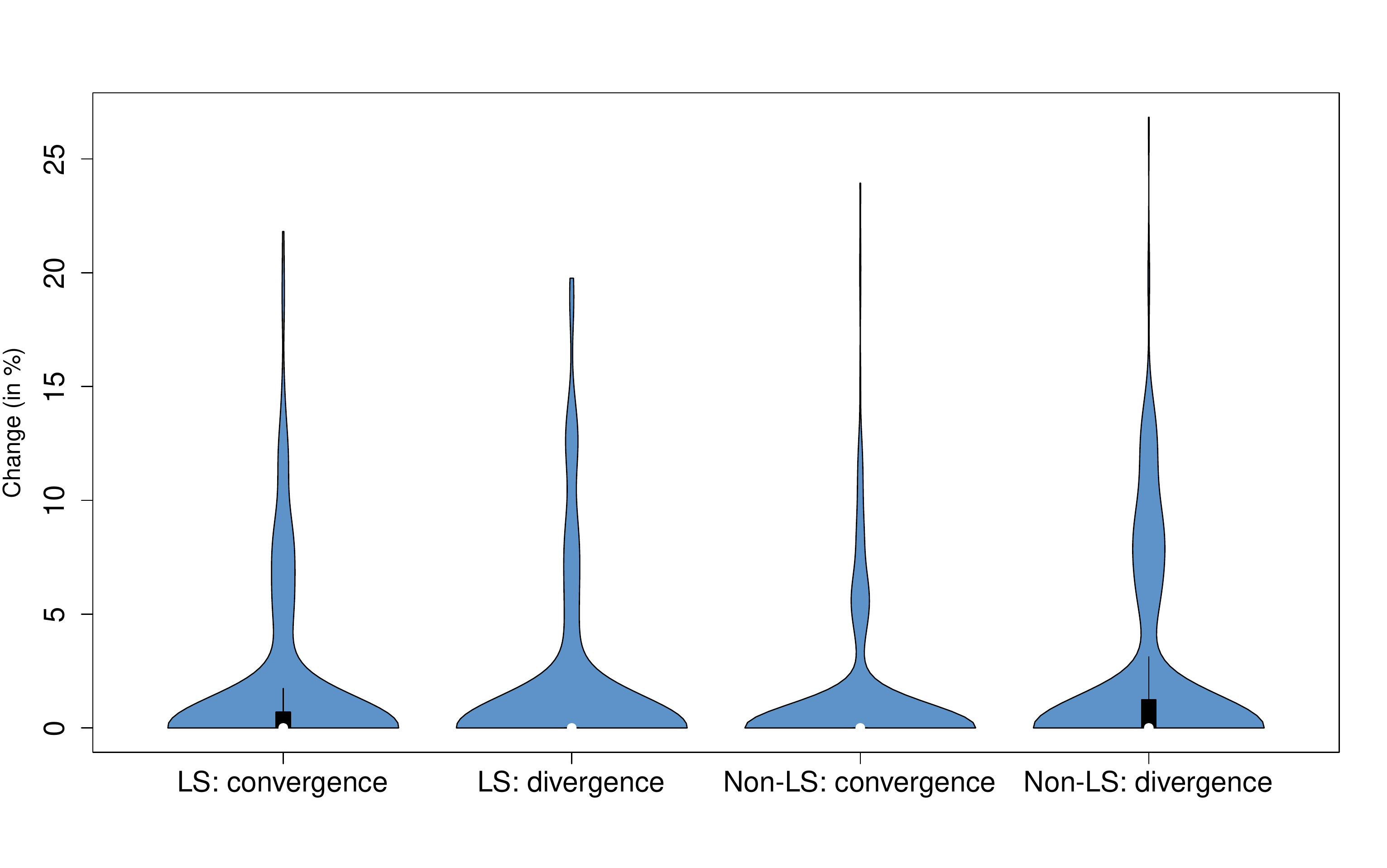}
	\caption{Violin plots showing the distributions of convergent and divergent change for Low Saxon (without Groningen; \textit{LS}) dialects and the other dialects (\textit{Non-LS}).}
	\label{fig:violin}
\end{figure}

When distinguishing the four different groups (FR, DU-FR, GR, LS), model comparison reveals that adding this predictor does not offer an improvement over a (null) model without the group distinction. However, on the basis of the geographical analysis showing more convergence in the north as opposed to the south, we also investigated whether including a contrast between Low Saxon (excluding Groningen) and the other three combined areas (i.e.~FR, DU-FR, and GR) resulted in an improved model. This was the case and the results on the basis of this optimal model (including the best random-effects structure) are shown in Tables \ref{tab:parametric_isLS} and \ref{tab:smooth_isLS}. Figure~\ref{fig:violin} visualizes these results. The final model specification is: \\ \vspace{1em}

\noindent 
\texttt{normalized\_change $\sim$ isLS * direction +  s(word, bs={\textquotesingle}re{\textquotesingle}) +\\ s(word, direction, bs={\textquotesingle}re{\textquotesingle})}\\ \vspace{0.5em}

\noindent
As before, birth year and age of the speakers did not improve the model, so we left these variables out of the model. The model explains 17.8\% of the variance in the data, which is roughly the same as the geographical model. While the geographical model was slightly better than the group-based model, this improvement was not significant ($p = 0.83$).

\begin{table}[!ht]
\caption{Parametric coefficients for a GAMM predicting change on the basis of a binary distinction between the Low Saxon group and the other dialect groups.}
\label{tab:parametric_isLS}
\begin{tabular}{@{}lrrrrr@{}}
\toprule
                                & \multicolumn{1}{l}{\textbf{Estimate}} & \multicolumn{1}{r}{\textbf{SE}} & \multicolumn{1}{l}{\textbf{$t$-value}} & \multicolumn{1}{l}{\textbf{$p$-value}} & \multicolumn{1}{l}{} \\ \midrule
Intercept (non-LS): convergence  & 0.013 & 0.003 & 4.154   & \textless{} 0.001 & ***\\
isLS: convergence               & 0.006 & 0.003 & 2.052   & 0.04 & *  \\
non-LS: divergence              & 0.010 & 0.004 & 2.553   & 0.01 & * \\
isLS: divergence                & -0.012   & 0.004 & -2.922  & \textless{} 0.01 & ** \\ \bottomrule
\end{tabular}
\end{table}

\begin{table}[!ht]
\caption{Smooth function terms (i.e.~random effects) for a GAMM predicting change on the basis of a binary distinction between the Low Saxon group and the other dialect groups.}
\label{tab:smooth_isLS}
\begin{tabular}{@{}lrrrr@{}}
\toprule
                     & \textbf{Edf} & \textbf{$F$-value} & \textbf{$p$-value} &     \\ \midrule
s(word) & 13.57 & 2.917 & \textless{}0.01          &    ** \\
s(word, direction) & 36.01 & 2.356 & \textless{}0.001          &    *** \\ \bottomrule
\end{tabular}
\end{table}

We can conclude from the model that convergence (towards standard Dutch) in the Low Saxon group is significantly higher, and divergence in the Low Saxon group is significantly lower than for the other groups. Moreover, divergence rates are significantly higher than convergence rates for the non-Low Saxon groups. We turn to this in the discussion. 

\section{Discussion}
The goal of this study was to investigate the amount of phonetic change in the north of the Netherlands using a real-time paradigm, while accounting for the direction of these changes as well. The results show that the overall level of phonetic change is low across a time span of approximately 20 years, at least for this geographical region and this particular time period. The percentages of change we obtained differ from \cite{heeringadialect2015}, who found an average percentage of phonetic change of 13.3\% across the whole Netherlandic area in an apparent time analysis (6.8\% convergence, 0.6\% neutral change, 5.9\% divergence), whereas we obtain on average only 3.4\% change (by summing the average convergence and divergence in our data, which are 2.0\% and 1.4\%, respectively). Several explanations are possible for this difference. Specifically, we investigated a much smaller area, and we used a different type of operation weights (PMI values tend to be much smaller than 1, even after normalization). Moreover, our comparison comprises approximately 22 years of real time language change, while Heeringa and Hinskens covered approximately 30 years of change in apparent time.\footnote{We repeated our analyses using the younger female data from DiaReg in order to validate our effects. In this combined real-time/apparent-time analysis, we find the same patterns of convergence within the Low Saxon area, but the effects are stronger, which is expected given the greater time span of language change (in a combination of apparent and real time).} 

Within the relatively small range of phonetic change we observed, the northern dialects show more divergence than convergence.  By contrast, when we look at the results for the more southern (non-Groningen) Low Saxon area, we see that there is consistently more convergence, which suggests that dialects that are geographically closer to the Randstad, the economic and cultural center of the Netherlands, seem to converge more towards Standard Dutch. These results are in line with synchronic results obtained by \cite{wielingquantitative2011}. 

The strong convergence towards Standard Dutch makes the Low Saxon dialect group a possible candidate for regiolectization if it coincides with significant convergence between neighboring dialects. However, to determine this, it would be necessary to analyze patterns of horizontal change together with the already investigated vertical change. In the future, it may therefore be interesting to investigate methods that can, in addition to vertical change, estimate horizontal change, for instance by extending the Levenshtein distance to even higher dimensions (cf.~\citealp{heeringadialect2015}). 

The comparative prominence of divergent over convergent patterns in the non-Low Saxon areas was a surprising finding of the group-based model. When we look at these dialect groups separately, we see that this pattern is caused by significant divergence rates in the Groningen and Dutch-Frisian dialects. Recall that the Dutch-Frisian dialects are perhaps best seen as comparatively stable language islands within the Frisian dialect area. We can confirm these notions in our data as well: the summed average rates of convergence to and divergence from Standard Dutch for these dialects (3.1\%) are slightly lower than those for the rest of the dialect continuum (3.5\%), though the convergence rates are much lower than the divergence rates. This suggests that the influence of Standard Dutch is marginal for these dialects, especially when we compare this to the Low Saxon varieties. We considered that these dialects may be changing towards a perceived `standard' variety within the Frisian group (which is the Grou dialect in DiaReg; cf.~\citealp{heeringa2005dialect}). It turned out this was not the case either, so the small changes in these dialects are not towards standard varieties. These patterns suggest that Dutch-Frisian dialects are either changing (1) horizontally (i.e.,~towards or away from neighboring varieties; cf.~the results of \citealp{heeringadialect2015}), or (2) internally (e.g.,~due to phonological processes that are not driven by language contact). 

For the Groningen dialects, there are examples in the literature that clearly indicate divergent patterns, such as the ongoing [ai] $\rightarrow$ [\textopeno i] change (\citealp{bloemhoffhandboek2008}, p. 162; observed in [stain(\textipa{@})] $\rightarrow$ [st\textopeno in(\textipa{@})] `stone' in the Dia\-Reg data for Groningen dialects). Note that this counts as divergence, because the PMI-based distance between [ai] and [ei] is smaller than between [\textopeno i] and [ei]. If we assume the Hollandic expansion (i.e.,~Hollandic speech norms expanding to outer provinces due to North and South Holland being the center of economic and political power; cf.~\citealp{kloeke1927hollandsche}) to be an ongoing process, it is possible that the Groningen dialects have so far resisted significant convergence towards Standard Dutch, simply due to their geographical distance from the Randstad. \citet{nerbonne2010measuring} found that dialect variation was (logarithmically) related to geographical distance in a large aggregated dataset, which offers some support to this tentative conclusion.

Frisian dialects form an exception to this tendency of divergence being greater than convergence, as there is no significant difference between convergence and divergence for this group. In other words, on an aggregate level the Frisian dialects remain approximately equidistant from Standard Dutch. This finding is in line with, e.g.,~\cite{buurke2020ma}, who showed relative stability of the Frisian dialects, and it is plausible given the protection this dialect group enjoys from its vital speaker population, the size of which incidentally also decreases less fast compared to other dialect groups \citep{driessen2012ontwikkelingen}. 

Note that both the geographical model and the model containing dialect distinctions (including the random-effects structure) only explained approximately 18\% of the data. Given that we have solely used linguistic and spatial information, this is not necessarily surprising. However, in future work, it would be useful to focus on collecting extralinguistic information and individual speaker preferences about their language use and attitudes towards dialects (their own and otherwise). Including this type of information will potentially increase the explanatory power of dialectometric studies (cf.~\citealp{wieling2015advances}). For example, we ideally would not only hypothesize that dialect vitality determines the rate and direction of change, but also operationalize the factors that are thought to underlie vitality (e.g.,~individual speaker attitudes and attitudes of the speaker population at large through questionnaires). The advanced mixed-effects regression models that are available nowadays, will allow for accounting for all these variables in a comprehensive model of language change (cf.~\citealp{wielingquantitative2011} for a synchronic example). 

We now consider some caveats of our data and analyses. The most pressing of these is perhaps the fact that in some locations relatively few words could be used for analyses, after our somewhat stringent inclusion criteria for the phonetic transcriptions. Our dataset comprised of a maximum of 36 words across 24 locations, while \cite{heeringadialect2015} used a maximum of 125 words for each location. However, preliminary simulations in our lab have shown that when using only 25 randomly selected words out of a larger set of hundreds of words, the results correlated at a level of $r = 0.9$ with the results of the larger dataset. Consequently, while our dataset is small, it does not appear too small. 

Upon closer inspection of the data from more divergent locations, we can observe that divergence may stem from the insertion or deletion of a single sound. This in itself is not unrealistic, because phonetic change can be very subtle. Examples (from Sneek and Grijpskerk) of such minimally different pairs include [kr\textipa{E}i\textipa{@}] $\rightarrow$ [kr\textipa{E}i\textipa{@}n] (Standard Dutch [kr\textipa{E}i\textgamma\textipa{@}(n)] `to get'), [la\textlengthmark t\textipa{@}r] $\rightarrow$ [la\textlengthmark t\textipa{@}] (Standard Dutch [la\textlengthmark t\textipa{@}r] `later'), and [lo\textlengthmark pt] $\rightarrow$ [lo\textlengthmark p] (Standard Dutch [lo\textlengthmark pt] `walks'), but also [d\oe\textlengthmark r] $\rightarrow$ [d\oe\textlengthmark ] (Standard Dutch [do\textlengthmark r] `through') and [l\textopeno\textlengthmark t\textipa{@}r] $\rightarrow$ [l\textopeno\textlengthmark  tr] (Standard Dutch [la\textlengthmark t\textipa{@}r] `later'). However given the two specific datasets we used, it is possible that these minor differences are a consequence of differences in task requirements. Recall that the DiaReg participants were asked to recount a story, whereas the GTRP participants translated isolated words and phrases. In running speech, the pragmatic pressure to communicate efficiently causes speakers to `simplify' their speech gestures (see \citealp{plug2006phonetic} for an overview), and we may be observing lenition in our data as well. For example, this may show as word-final [n, t, r] not being pronounced or the realization of [\textipa{@}] being so minimal that it cannot be heard by transcribers. Ideally, the phonetic data should be more comparable in this regard to avoid this uncertainty. Consequently, we intend to collect an additional dataset in the near future using a methodology closer akin to that of the GTRP, for which we ensure a large overlap with both the GTRP and the DiaReg. For the present study, however, we are not able to draw any firm conclusions about the extent to which these dataset differences affect our results, simply because we cannot adequately verify whether participants reduced their speech.  

One more general shortcoming concerns our artificial partitioning of language change into phonetic change and change on different linguistic levels, including simultaneous change at the lexical and phonetic levels due to lexical borrowing. This process is cross-linguistically widespread in contact situations \citep{taylor_lexical_2015} and the possible interplay between phonetic and lexical change is perhaps more important than we give it credit for in this study. The longstanding language contact between, for example, the Low Saxon and Low Franconian areas has led to considerable lexical borrowing between the corresponding languages (though mostly unilaterally from Standard Dutch into Low Saxon). However, a large amount of lexical borrowing does not entail an equal amount of phonetic change. For example, the traditional Groningen dialect word for ice-skating was \textit{scheuvelen} [s\textipa{x}\o\textsubring{v}\textipa{@}\textltilde\textsyllabic{n}], but nowadays it is not uncommon to hear \textit{schoatsen} [s\textipa{x}\textupsilon\textlengthmark ts\textsyllabic{n}] instead, which is borrowed from Dutch \textit{schaatsen} [s\textipa{x}a\textlengthmark ts\textipa{@}(n)], but also stays true to the regular correspondences between standard Dutch and Groningen dialects (i.e.,~[a\textlengthmark ] in Standard Dutch cognates being pronounced as [\textupsilon\textlengthmark ]). This illustrates how the phonetic level may be more resistant to change than the lexical level, which is in line with \cite{heeringadialect2015}. 

Finally, a point should be made about approximating a standard variety. We mentioned in passing that news presenters are typically considered to be representative speakers of the standard variety. This is certainly the case, but we have only used the transcriptions from a single news presenter. We can improve upon this approach in two ways. The obvious shortcoming is that even highly experienced national news presenters with an intended wide coverage are still not accentless, not even when they are told to be as neutral as possible in an experimental setting. They may therefore not always be perceived as standard speakers by individual listeners. This is only natural, but since we are only using a single speaker as a reference point, this does make it possible that our results are skewed towards this idiomatic version of the standard variety. A second reason why the use of a single reference point is problematic is that standard varieties change as well. The Standard Dutch that is spoken nowadays is noticeably different from the standard variety 50 years ago \citep{smakman2006standard}, so ideally, reference speakers should be chosen that represent the standard from around the average recording year for each dataset. Both these problems can be solved in future work by using data from more than one standard Dutch speaker and selecting standard Dutch speakers from different age ranges.

\section{Conclusion}
In this study, we found, using sophisticated dialectometric analyses, that phonetic change progresses slowly in the northern part of the Netherlandic dialect continuum. There was much more stability than there was change in our data, but we observed different patterns between dialect groups within the ongoing changes. Areas that are further away from the economic center of the Netherlands, or are otherwise protected by their vital speaker populations, seem to escape the ongoing expansion of Standard Dutch. The Low Saxon areas (excluding Groningen) benefit from neither of these protective measures, and therefore noticeably converged to Standard Dutch even across a relatively short time period.

\bibliography{bibliography}
\bibliographystyle{newapa}

\end{document}